\documentclass[10pt,twocolumn,letterpaper]{article}

\usepackage{cvpr}              
\definecolor{cvprblue}{rgb}{0.21,0.49,0.74}
\usepackage[pagebackref,breaklinks,colorlinks,allcolors=cvprblue]{hyperref}

\usepackage{multirow}
\usepackage[table]{xcolor}
\usepackage{pifont}
\usepackage{caption}

\usepackage{listings}
\usepackage{xcolor}
\definecolor{codeblue}{rgb}{0.25,0.5,0.75}
\definecolor{codegray}{rgb}{0.5,0.5,0.5}
\definecolor{codepurple}{rgb}{0.58,0,0.82}
\definecolor{codepink}{rgb}{0.8, 0, 0.45}

\lstdefinestyle{mypython}{
  language=Python,
  basicstyle=\ttfamily\small,
  keywordstyle=\color{codepink}\bfseries,
  commentstyle=\color{codegray}\itshape,
  stringstyle=\color{codepurple},
  showstringspaces=false,
  breaklines=true,
  tabsize=2,
  frame=none,
  numbers=left,
  numberstyle=\small\color{codegray},
  numbersep=6pt,
  xleftmargin=-1pt,
  alsoletter=+-*/=<>!,
  moredelim=[is][\color{codepink}]{@@}{@@},
  literate=
    {.}{{\textbf{.}}}1
    {+}{{{\color{codepink}+}}}1
    {-}{{{\color{codepink}-}}}1
    {*}{{{\color{codepink}*}}}1
    {/}{{{\color{codepink}/}}}1
    {=}{{{\color{codepink}=}}}1
    {>}{{{\color{codepink}>}}}1
    {<}{{{\color{codepink}<}}}1
    {!}{{{\color{codepink}!}}}1
    {:}{{{\color{codepink}:}}}1
    {@}{{{\color{codepink}@}}}1
}

\lstnewenvironment{PythonB}[1][]{
    \lstset{style=mypython,#1}
}{}

\usepackage[ruled,boxed,linesnumbered]{algorithm2e}
\usepackage{orcidlink}
\usepackage[normalem]{ulem}

\newcommand{\cmark}{\ding{51}}%
\newcommand{\xmark}{\ding{55}}%

\newcommand{\specialcell}[2][c]{%
  \begin{tabular}[#1]{@{}c@{}}#2\end{tabular}}

\newcommand{\modelname}{{SAVE}\xspace}

\newcommand{\cradd}[1]{\textcolor{black}{#1}}

\newcommand{\drop}[1]{\textcolor{red}{#1}}

\def\paperID{} 
\def\confName{CVPR}
\def\confYear{2026}

\title{SAVE: \underline{S}peech-\underline{A}ware \underline{V}ideo R\underline{e}presentation Learning for Video-Text Retrieval}

\author{Ruixiang Zhao\thanks{Equal contributions.} \orcidlink{0009-0008-9984-1841}
\hspace{6mm}Zhihao Xu$^{*}$\hspace{6mm}Bangxiang Lan\hspace{6mm}Zijie Xin\hspace{6mm}Jingyu Liu\hspace{6mm}Xirong Li\thanks{Corresponding author: Xirong Li (xirong@ruc.edu.cn)} \orcidlink{0000-0002-0220-8310}\\
Renmin University of China\\
\tt\small\href{https://github.com/ruc-aimc-lab/SAVE}{https://github.com/ruc-aimc-lab/SAVE}
}

\begin{document}
\maketitle
\begin{abstract}
For video-text retrieval, the use of CLIP has been a \emph{de facto} choice. Since CLIP provides only image and text encoders, this consensus has led to a biased paradigm that entirely ignores the sound track of videos. While several attempts have been made to reintroduce audio -- typically by incorporating an audio encoder and fusing its output with visual features -- these methods face two challenges: \textbf{ineffective representation of speech content} and \textbf{suboptimal vision-audio fusion}. To address these issues jointly, we propose \textbf{SAVE}, a \textbf{S}peech \textbf{A}ware \textbf{V}ideo r\textbf{E}presentation learning  method. SAVE improves upon AVIGATE, a SOTA audiovisual method, with a dedicated speech branch for more effective speech embedding. Furthermore, we introduce soft-ALBEF for early vision-audio alignment that facilitates fusion. Extensive experiments on five benchmarks show that SAVE compares favorably against the SOTA, outperforming AVIGATE by +4.1\% on \cradd{MSRVTT-9k}, +1.9\% on \cradd{MSRVTT-7k}, +2.5\% on VATEX, +9.8\% on Charades, and +2.1\% on LSMDC, in light of the SumR metric. 
\end{abstract}    
\section{Introduction}
\label{sec:intro}

\begin{figure}[ht!]
    \centering
    \begin{subfigure}[b]{\linewidth}
        \centering
        \includegraphics[width=\linewidth]{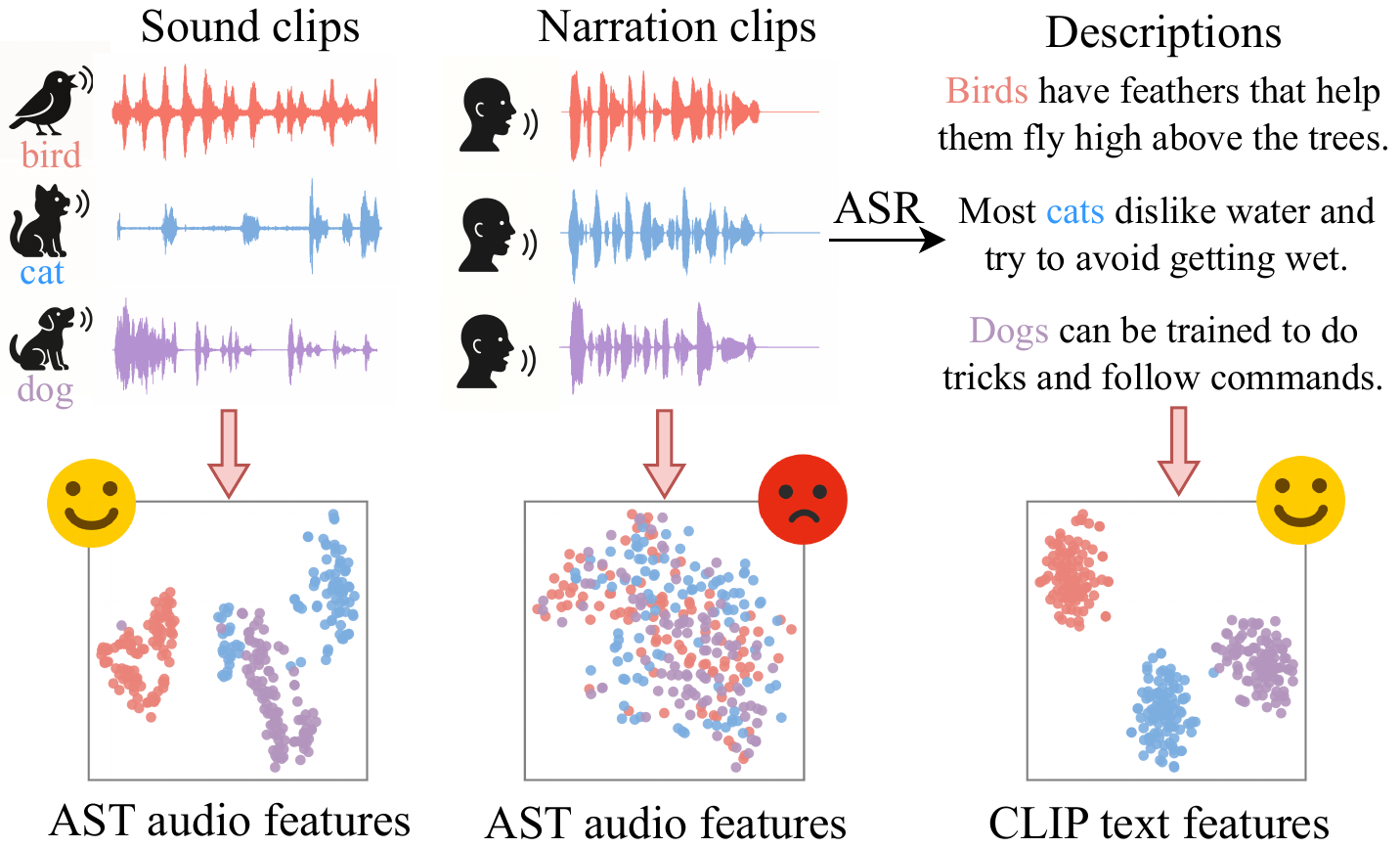}
        \caption{Limitation of current audio encoders}
    \label{fig:first_a}
    \end{subfigure}

    \vspace{4mm}
    \begin{subfigure}[b]{\linewidth}
        \centering
        \includegraphics[width=\linewidth]{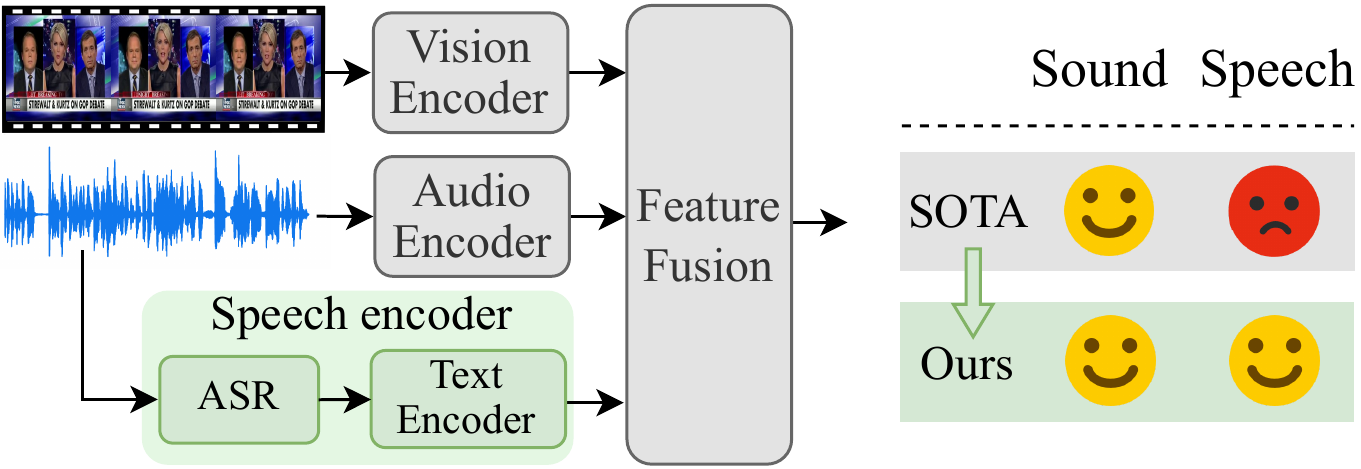}
        \caption{Proposed SAVE method for speech-aware video representation learning}
    \label{fig:first_b}
    \end{subfigure}
    
    \vspace{4mm}
    \begin{subfigure}[b]{\linewidth}
        \centering
        \includegraphics[width=\linewidth]{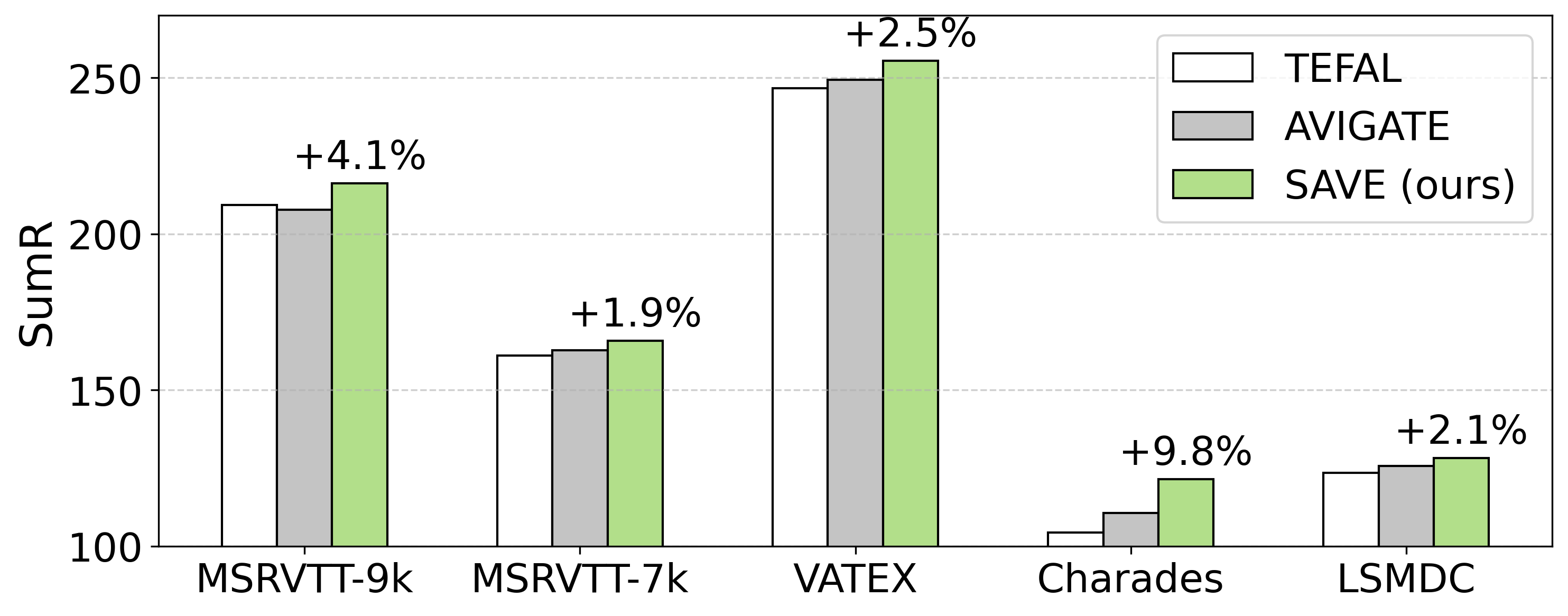}
        \caption{Comparing with SOTA audio-enhanced models}
    \label{fig:first_c}
    \end{subfigure}

    \caption{\textbf{An overview of this paper}. \textbf{(a) Problem}: 
    Current audio encoders (\cradd{ResNet-18} \cite{resnet} and AST \cite{ast}), trained on datasets of environmental sounds, are not well suited for speech embedding.
    \textbf{(b) Solution}: We improve the state-of-the-art in audio-enhanced video-text retrieval by introducing a dedicated speech branch for speech-aware video embedding. 
    \textbf{(c) Results}: Our method consistently outperforms SOTA audio-enhanced models (TEFAL \cite{tefal} and AVIGATE \cite{avigate}) across five benchmarks.}
    \label{fig:first}
\end{figure}

Video-text retrieval (VTR), a fundamental task in vision-language understanding, aims to retrieve the most semantically relevant video \wrt a given text query or vice versa. Owing to the remarkable success of Contrastive Language-Image Pre-training (CLIP) \cite{clip} in bridging visual and textual modalities, the use of CLIP has been a \emph{de facto} choice for VTR \cite{clip4clip,clipvip,xpool,teachclip,pig}. Since CLIP provides only image and text encoders, these CLIP-based methods naturally \textbf{neglect the sound track} of videos.

In order to reintroduce the sound track, several attempts have been made \cite{eclipse, tefal, avigate}, see \cref{tab:works}. They typically place an audio encoder alongside the image encoder for audio embedding. Subsequently, audio-enhanced video embedding is obtained by fusing the audio embedding with its visual counterpart from the vision encoder. Concerning the choice of vision-audio fusion, text-conditioned attention is considered in TEFAL \cite{tefal}, whilst symmetric and asymmetric cross-attention based fusions are used in EclipSE \cite{eclipse} and AVIGATE \cite{avigate}, respectively. 
These audiovisual methods have demonstrated superior performance compared to their vision-only counterparts.

\begin{table}[!tbp]
\caption{\textbf{SOTA audiovisual methods for video-text retrieval}. }
\centering
\setlength{\tabcolsep}{3pt} 
\renewcommand{\arraystretch}{1.1} 
\resizebox{\linewidth}{!}{
\begin{tabular}{@{}llcccr@{}}
\toprule
\textbf{Method}  & \textbf{Audio} & \textbf{Speech} & \specialcell{\textbf{Video}\\ \textbf{embedding}\\ \textbf{network}} & \specialcell{\textbf{Early}\\ \textbf{vision-audio}\\ \textbf{alignment}} & \specialcell{\textbf{\cradd{MV-9k}}\\ \textbf{SumR}} \\
\midrule
EclipSE\cite{eclipse}  & \cradd{ResNet-18} & \xmark & Bi-branch & \xmark & 197.8 \\
TEFAL\cite{tefal}   & AST & \xmark & Bi-branch & \xmark & 209.2 \\
AVIGATE\cite{avigate}  & AST & \xmark & Bi-branch & \xmark & 207.7 \\
\rowcolor{green!20}
\emph{SAVE} (ours)  & AST & ASR + CLIP & Tri-branch & \cmark & \textbf{216.2} \\
\bottomrule
\end{tabular}
}
\label{tab:works}
\end{table}

With a closer look at the audio encoders currently used, \ie \cradd{ResNet-18} \cite{resnet} in EclipSE and AST \cite{ast} in TEFAL and AVIGATE, we notice that they are not well suited for speech embedding. The reason is surprisingly simple: the audio encoders were not trained for speech content understanding. To demonstrate such a deficiency, \cradd{we design a toy experiment to inspect whether existing audio encoders can produce well-separated embeddings for speech content.} For three animal classes, \ie \emph{bird}, \emph{cat}, and \emph{dog}, we construct three modality-specific datasets:  (i) \textit{Sound}: Real animal vocalizations collected from AudioSet \cite{Audioset}, (ii) \textit{Speech}: Narration clips\footnote{We first prompt an LLM to generate text descriptions about the characteristics per animal, and then use a Text-to-Speech model to convert these texts into narration clips.} describing characteristics per animal, and (iii) \textit{Text}: Transcripts obtained by performing Automatic Speech Recognition (ASR) with Whisper \cite{whisper} on the \emph{Speech} dataset. Each dataset contains 100 samples per class. Sound and speech embeddings are acquired with AST, while text embeddings are produced by the CLIP text encoder. 
As shown in \cref{fig:first_a}, while the sound and text embeddings are well separated class-wisely, the \textbf{speech samples} are \textbf{cluttered in} the \textbf{audio feature space}\footnote{Similar cluttered results are also observed in the audio feature space of Whisper, see the supplement, though it has good ASR performance.}.

\begin{figure}[!htbp]
\centering
\includegraphics[width=\columnwidth]{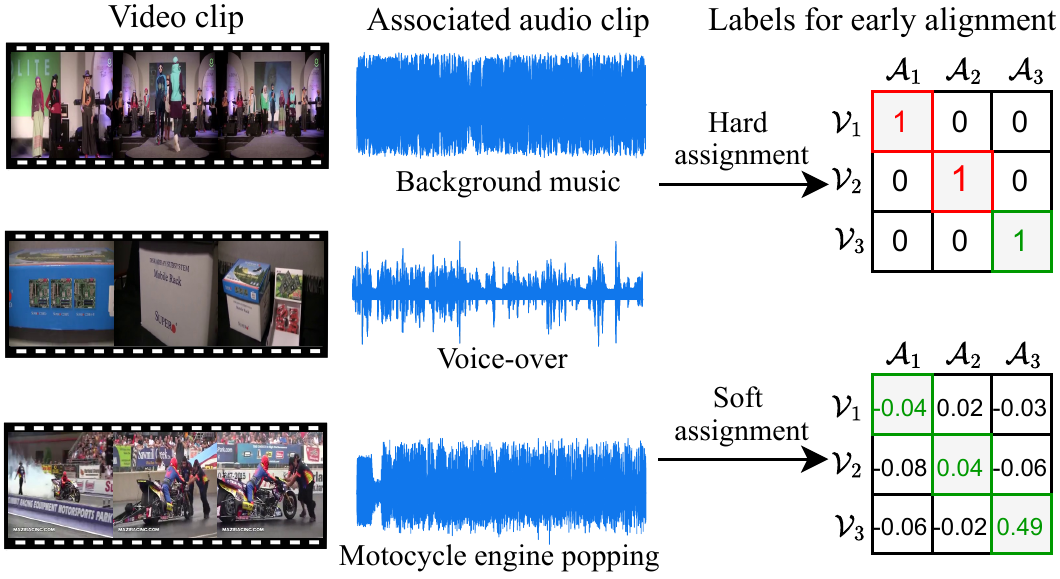}
\caption{\textbf{Hard \emph{vs}. soft labels for early vision-audio alignment}. For the videos in the first two rows, their associated sound tracks are not semantically relevant \wrt the video content. Enforcing vision-audio alignment for these videos is adverse. By contrast, the soft labels, estimated by ImageBind \cite{imagebind}, provide finer supervision for better vision-audio alignment.}
\label{fig:soft_label}
\end{figure}

Another overlooked issue in the current audiovisual methods is the difficulty in vision-audio fusion. Consider AVIGATE for instance. The method fully relies on a cross-attention based module to directly fuse the frame features (produced by the CLIP image encoder) and the audio features (produced by AST), which are misaligned by definition. Meanwhile, lessons learned in the context of vision-language pre-training show that \emph{align before fuse} (ALBEF) \cite{albef} helps. That said, directly applying ALBEF for the vision-audio alignment is problematic. Unlike image-caption pairs used for vision-language pre-training, vision-audio pairs often lack semantic correspondence, see \cref{fig:soft_label}. Consequently, enforcing \textbf{early vision-audio alignment} on the noisy pairs \textbf{with hard labels} may cause the model to \textbf{learn spurious correlations} rather than genuinely meaningful alignment.

In order to jointly address the two issues, we propose in this paper a \textbf{S}peech \textbf{A}ware \textbf{V}ideo r\textbf{E}presentation learning  method (\textbf{SAVE}). More specifically, for effective speech embedding, we extend the AVIGATE framework with a dedicated speech branch. Meanwhile, we develop soft-ALBEF for early vision-audio alignment, harnessing ImageBind \cite{imagebind} to provide soft, noise-tolerant supervision. 
To sum up, our main contributions are three-fold:  \\
$\bullet$ \textbf{The first speech-aware video embedding for CLIP-based VTR}. We introduce a dedicated speech branch that converts narration audio into text via ASR and encodes it with a text encoder, enabling CLIP-based retrieval models to explicitly capture semantic speech information that traditional audio encoders fail to represent. \\
$\bullet$ \textbf{Soft-ALBEF for robust early vision-audio alignment}.
We propose a noise-tolerant alignment strategy that leverages ImageBind to generate soft supervision signals, effectively mitigating the semantic mismatch and spurious correlations caused by directly enforcing hard alignment on noisy vision-audio pairs. \\
$\bullet$ \textbf{New state-of-the-art}. SAVE achieves superior performance across five VTR benchmarks, outperforming AVIGATE by +4.1\% on \cradd{MSRVTT-9k}, +1.9\% on \cradd{MSRVTT-7k}, +2.5\% on VATEX, +9.8\% on Charades, and +2.1\% on LSMDC, in light of the SumR metric.

\section{Related Work}
\label{sec:work}

This work aims to learn speech-aware video representation for VTR. Accordingly, we review three relevant lines of research: CLIP-based VTR, audio-enhanced VTR, and the use of speech for VTR.

\textbf{CLIP-based VTR}. 
CLIP4Clip \cite{clip4clip} is the first to adopt CLIP for end-to-end VTR. More recent models improve the retrieval performance by introducing fine-grained interactions between the vision and text modalities, including frame-sentence alignment \cite{uatvr,xpool,drl,eercf}, frame-word alignment \cite{hbi,xclip,videocolbert,bima}, and patch-word alignment \cite{pidro,prost,ucofia}. Another direction of research on CLIP-based VTR emphasizes efficiency in training / inference. For efficient training, VoP \cite{vop}, MV-Adapter \cite{mvadapter} and DGL \cite{dgl} resort to parameter-efficient fine-tuning. For inference efficiency, existing efforts can be roughly divided into two groups. The first group uses knowledge distillation to transfer knowledge from advanced yet computationally heavy teacher models to student models that are computationally more efficient \cite{clipping,teachclip}. The second group is  based on token compression \cite{centerclip,testa,tempme},  reducing memory footprint and computational cost by generating compact yet expressive tokens for video-text matching. The above methods do not consider the sound track.

\textbf{Audio-enhanced VTR}. 
Methods for audio-enhanced VTR put an audio encoder alongside the CLIP vision encoder \cite{eclipse,tefal,avigate}. EclipSE \cite{eclipse} employs \cradd{ResNet-18} \cite{resnet} as its audio encoder, with a symmetrical cross-attention mechanism to fuse features from the visual and audio branches. TEFAL \cite{tefal} and AVIGATE \cite{avigate} use Audio Spectrogram Transformer (AST) \cite{ast} as a stronger audio encoder. \cradd{TEFAL adopts X-Pool \cite{xpool} for both text-audio alignment and text-vision alignment. Such a dual use of X-Pool creates a large computational overhead. AVIGATE does not conduct text-audio alignment. Instead, the method integrates the audio feature into the video feature by cross-attention based gated fusion. }  
Despite its state-of-the-art performance, AVIGATE has two drawbacks. 
First, current audio encoders, primarily trained for modeling environmental sounds, are ineffective in capturing speech semantics. As a consequence, the video representation of AVIGATE is not speech-aware. 
Second, unlike the text and vision features, which are already pre-aligned by CLIP, the audio and vision features have not undergone such pre-alignment. The lack of pre-alignment hinders the effective fusion of audio and vision features.

\textbf{The Use of Speech for VTR}. 
Before the advent of CLIP-based methods, some initial efforts were made to utilize ASR-generated text for VTR \cite{schmidt2011two, yang2014content,ce,mmt,maskmmt}. In \cite{schmidt2011two, yang2014content}, text-to-video retrieval is implemented by matching a specific query with the ASR text. Later studies, such as CE \cite{ce}, MMT \cite{mmt}, and Masked-MMT \cite{maskmmt}, simply treat the ASR text, vectorized by a pre-trained Word2Vec model, as one of the many video features to be fused.  Such an ASR feature is found to be weak, practically overlooked during the feature fusion process. 

\cradd{Our work builds upon AVIGATE and addresses its drawbacks as follows}. We add a third branch to the AVIGATE network for explicit speech content representation. Inspired by the successful use of ASR text in cross-domain product retrieval \cite{tmm26-zhao}, we choose to encode the speech modality using CLIP's text encoder.
We develop soft-ALBEF for pre-alignment of the audio and vision features before fusion. These designs lead to clear improvements over the SOTA audiovisual VTR baselines (\cref{tab:works}).

\begin{figure*}[!htbp]
  \centering
  \includegraphics[width=0.95\linewidth]{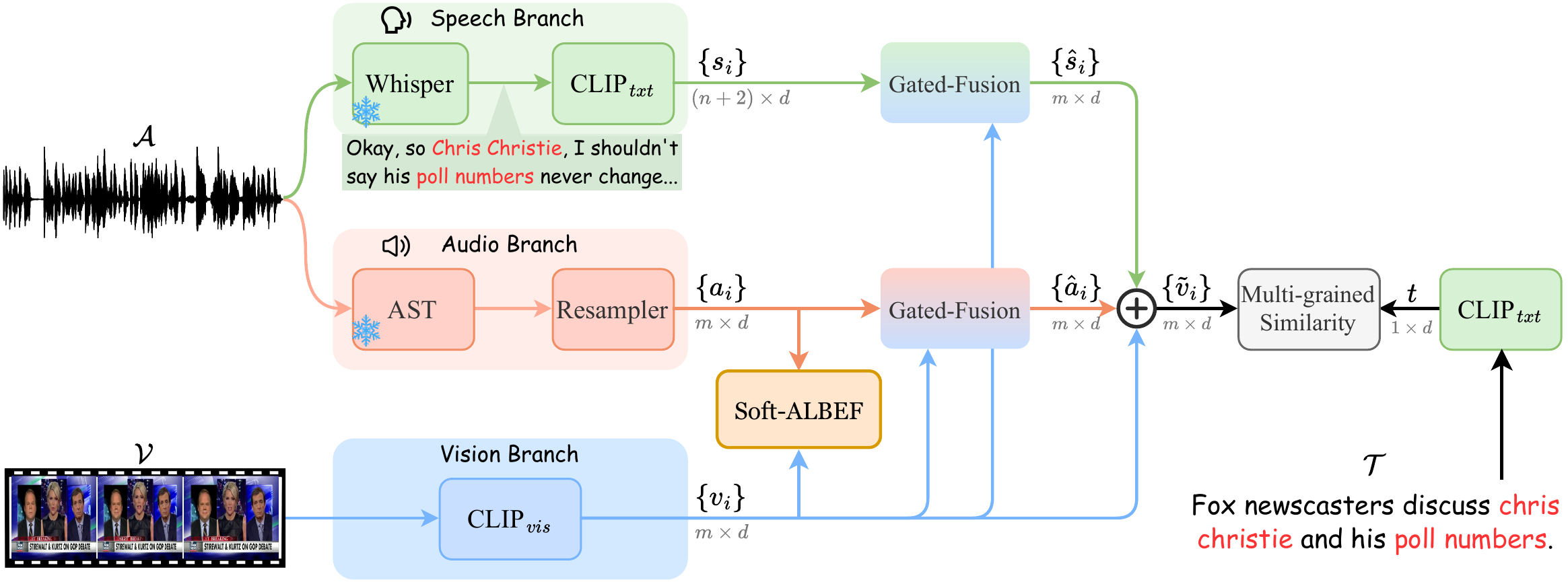}
  \caption{\textbf{Proposed \underline{s}peech-\underline{a}ware \underline{v}ideo r\underline{e}presentation learning (SAVE) method for video-text retrieval}. Given a short video $\mathcal{V}$ associated with a sound track $\mathcal{A}$, \modelname uses a tri-branch network to embed the video frames to a set of visual tokens $\{v_i\}$,  the sound to a set of audio tokens $\{a_i\}$, and the speech to a set of textual tokens $\{s_i\}$. Gated fusion conditioned on the visual tokens is performed on the audio and textual tokens, yielding fused tokens $\{\hat{a}_i\}$ and $\{\hat{s}_i\}$, respectively. By aggregating  $\{v_i\}$, $\{\hat{a}_i\}$ and  $\{\hat{s}_i\}$, we obtain $\{\hat{v}_i\}$ as a speech-aware video representation. The video's relevance \wrt a specific query $\mathcal{T}$ is computed as a multi-grained similarity between $\{\hat{v}_i\}$ and the query embedding $t$.  Soft-ALBEF is used only during training for better alignment between the visual and audio tokens. 
  }
  \label{fig:framework}
\end{figure*}

\section{Method}\label{sec:method}

Given a video $\mathcal{V}$ with its soundtrack $\mathcal{A}$, we perform speech-aware video embedding via a tri-branch network. As \cref{fig:framework} shows, the network consists of a vision branch for  frame embedding, an audio branch for sound embedding, and a speech branch for spoken language embedding. For the vision and audio branches, we follow AVIGATE \cite{avigate}, the SOTA for learning audio-enhanced video representation. So in what follows, we first outline the AVIGATE framework  (\cref{ssec:avigate}), followed by the extension of the framework to the speech branch (\cref{ssec:speech}). We then introduce our soft-ALBEF strategy for early vision-audio embedding alignment and a joint loss for network training (\cref{ssec:soft-albef}). 

\subsection{AVIGATE in a Nutshell} \label{ssec:avigate}

We summarize the video embedding module of AVIGATE in \cref{eq:avigate}. For visual token extraction, an array of $m$ frames are uniformly sampled from the given video $\mathcal{V}$. Per frame, the visual encoder of the CLIP model \cite{clip} ($\mbox{CLIP}_{vis}$) is used to extract a $d$-dimensional frame embedding $v$, which is the  [CLS] token of the encoder's last layer. Processing all the $m$ frames results in a sequence of $m$ visual tokens $\{v_i\}$.

\begin{equation} \label{eq:avigate}
\left\{
\begin{array}{ll}
\{f_1, \ldots, f_m\} & \leftarrow \mbox{video-to-frames}(\mathcal{V}, m), \\
\{v_1, \ldots, v_m\} & \leftarrow \mbox{CLIP}_{vis}(\{f_1, \ldots, f_m\}), \\
\{a_1, \ldots, a_m\} & \leftarrow \mbox{Resampler}(\mbox{AST}(\mathcal{A}), m), \\
\{\hat{a}_1, \ldots, \hat{a}_m\} & \leftarrow \mbox{Gated-Fusion}(\{v_i\}, \{a_i\}), \\
\{\hat{v}_i\} & \leftarrow \cradd{0.95\{v_i\} + 0.05\{\hat{a}_i\}}. \\
\end{array}
\right.
\end{equation}

For audio token extraction,  the source track $\mathcal{A}$ is first transformed into Mel filter bank features and then fed into the AST model \cite{ast}, yielding a sequence of audio tokens. For token reduction, the sequence goes through a transformer-based resampler with $m$ learnable queries, generating a shortened sequence of $m$ $d$-dimensional audio tokens $\{a_i\}$. Next, a Transformer-based gated fusion is performed, with $\{v_i\}$ as $Q$ and $\{a_i\}$ as $K$ / $V$, to extract visually related audio tokens $\{\hat{a}_i\}$. A weighted sum of $\{\hat{a}_i\}$ and $\{v_i\}$ leads to $\{\hat{v}_i\}$ as audiovisual tokens.
The given video is jointly represented by $\{\hat{v}_i\}$ and their mean $\bar{v}$.

For a specific text $\mathcal{T}$ to be matched with $\mathcal{V}$, its $d$-dimensional embedding $t$ is obtained by the CLIP's text encoder ($\mbox{CLIP}_{txt}$). A global-level video-text similarity $s_g(\mathcal{V}, \mathcal{T})$ is computed as the cosine similarity between $\bar{v}$ and $t$. A local-level similarity $s_l(\mathcal{V}, \mathcal{T})$ is obtained by first computing the cosine similarity between $\hat{v}_i$ and $t$ and then aggregating the multiple scores using a log-sum-exp function. The final similarity is obtained as $(s_g + s_l)/2$.

\subsection{Adding a Speech Branch to AVIGATE} \label{ssec:speech}

In order to make the video representation speech-aware, we extend the AVIGATE network with a new speech branch. As shown in \cref{fig:framework}, we use Whisper large-v3 \cite{whisper},  a state-of-the-art ASR model, to transcribe the speech within the sound track to a sequence of $n$ words $\{w_1, \ldots, w_n\}$.  We then use  $\mbox{CLIP}_{txt}$ for textual token extraction, obtaining a sequence of $n+2$ $d$-dimensional tokens as $\{s_S, s_1, \ldots, s_n, s_E\}$, where $s_S$ and $s_E$ are the start and end tokens automatically added by the text encoder. 

We also use a Gated-Fusion module, with the visual tokens $\{v_i\}$ as $Q$ and the speech-derived tokens as $K$ and $V$, producing visually selected speech tokens $\{\hat{s}_1, \ldots, \hat{s}_m\}$. The key dataflow of our speech branch is summarized as
\begin{equation} \label{eq:save}
\left\{
\begin{array}{ll}
\{w_1, \ldots, w_n\} & \leftarrow \mbox{Whisper}(\mathcal{A}), \\
\{s_S, s_1, \ldots, s_n, s_E\} & \leftarrow \mbox{CLIP}_{txt}(\{w_1, \ldots, w_n\}), \\
\{\hat{s}_1, \ldots, \hat{s}_m\} & \leftarrow \mbox{Gate-Fusion}(\{v_i\}, \{s_i\}). \\ 
\end{array}
\right.
\end{equation}
Consequently, we obtain speech-aware audiovisual tokens $\{\tilde{v}_i\}$  as $\{v_i\} + (\{\hat{a}_i\} + \{\hat{s}_i\})/2$. \cradd{Our motivation for combining the modality embeddings in this manner is three-fold. First, lacking prior knowledge of videos to be retrieved, we treat the speech and the audio branches as equally important. Second, as the visual content generally matters more, averaging the speech and audio embeddings first and then combining them with the vision embedding effectively assigns a larger weight to vision in a parameter-free manner. Third, the simple fusion encourages the Gate-Fusion modules to learn which signals truly matter.}

\subsection{Training with Soft-ALBEF}\label{ssec:soft-albef}

\textbf{Soft Align before Fuse}. 
As formalized in \cref{eq:avigate,eq:save}, the Gated-Fusion modules play a crucial role in vision-audio and vision-speech feature interaction and fusion. For cross-attention based fusion to be effective, the query ($Q$) needs to be well aligned with the key ($K$). This alignment is naturally present for vision-speech fusion, where the visual tokens ${v_i}$ and speech tokens ${s_i}$ are pre-aligned through the shared CLIP space ($\mbox{CLIP}_{vis}$ and $\mbox{CLIP}_{txt}$). By contrast, such pre-alignment is absent for the vision-audio counterpart. Furthermore, the inherent noise in vision-sound pairs (\cref{fig:soft_label}) makes directly applying ALBEF \cite{albef} problematic. To address both issues, we replace ALBEF's hard labels with vision-sound relevance scores computed by ImageBind \cite{imagebind}, a strategy we term soft-ALBEF.

Given a batch of $B$ video-audio pairs $\{(\mathcal{V}_i, \mathcal{A}_i)\}_{i=1}^B$, we employ ImageBind to obtain a $B\times B$ video-audio affinity matrix $M_0$, with $M_0[i,j]$ indicating the cosine similarity between the ImageBind embeddings of the $i$-th video and $j$-th audio. Similarly, we compute a video-audio affinity matrix  $M_1$ from the pre-fusion video and audio embeddings generated by the current network\footnote{The video and audio embeddings are obtained by mean pooling over $\{v\}$ and $\{a\}$, respectively.}. We use $M_0$ as soft supervision to guide the generation of $M_1$ through a Pearson distance loss \cite{pearson}, \cradd{namely
\begin{equation} \label{eq:pearson}
\begin{array}{ll}
 \ell_{pearson} :=&\dfrac{1}{b} \sum\limits_{i=1}^b d_p\big(\sigma(M_0[i,\cdot]), \sigma(M_1[i,\cdot])\big) + \\ 
 & \dfrac{1}{b} \sum\limits_{j=1}^b d_p \big(\sigma(M_0[\cdot, j]), \sigma(M_1[\cdot, j]) \big)
\end{array}
\end{equation}
where $\sigma$ is softmax function, $d_p$ is the Pearson distance.}
This choice
is motivated by its invariance to separate changes in scale and location \cite{pearson, teachclip}, when compared to alternatives such as MSE or Huber loss. This property ensures that our network focuses on learning the relative ranking structure, 
rather than fitting absolute values, and is thus 
more robust and aligns better with our objective of capturing meaningful video-audio correspondences. 

\textbf{Loss}. The Pearson distance loss 
serves as an auxiliary objective to 
the primary adaptive margin-based contrastive loss from AVIGATE. The two losses are equally combined.

\begin{figure}[ht!]
\centering
\includegraphics[width=\columnwidth]{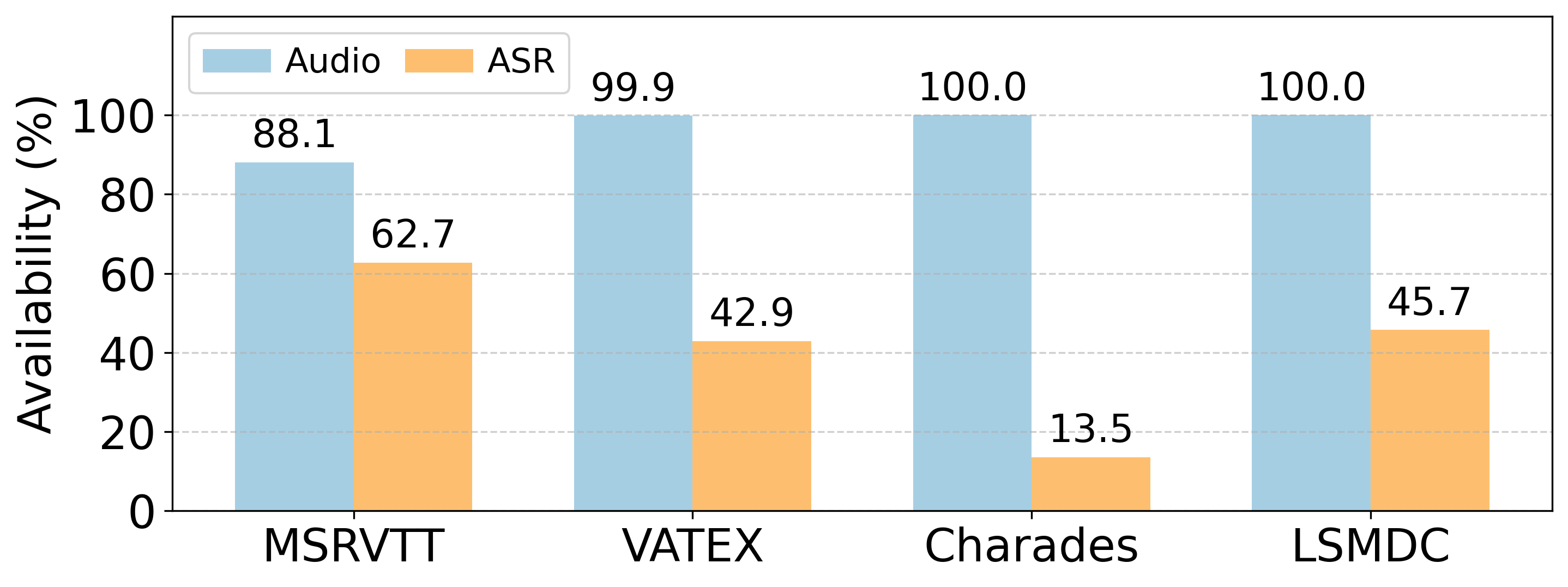}
\caption{\textbf{Proportion of videos with audio / ASR available}.}
\label{fig:dataset}
\end{figure}

\textbf{Handling Missing Data}. 
In practice, not all videos have sound tracks or ASR text available (\cref{fig:dataset}). When no sound track is present, the Mel filter bank is set to zero. For cases where ASR recognition fails, an empty string is used, which is then padded as a zero vector by the tokenizer of $\mbox{CLIP}_{txt}$.

\section{Experiments} \label{sec:experiments}

\subsection{Experimental Setup}
\label{ssec:setup}
\textbf{Datasets}. We evaluate our approach on several public datasets commonly adopted for audiovisual video-text retrieval research, including MSRVTT \cite{msrvtt}, VATEX \cite{vatex}, Charades \cite{charades}, and LSMDC \cite{lsmdc}.
For MSRVTT, we consider both the original \cradd{MSRVTT-7k} split and the \cradd{MSRVTT-9k} partition \cite{msrvtt-1ka}. 
For VATEX, we follow the partition introduced by Chen \etal \cite{hrg}. Regarding Charades and LSMDC, we use their official data splits. A summary of dataset statistics is given in \cref{tab:dataset}. Both \cradd{MSRVTT-9k} and \cradd{MSRVTT-7k} are used for the ablation study.

\begin{table}[htbp]
\centering
\caption{\textbf{Number of videos per benchmark}.} 
\setlength{\tabcolsep}{6pt} 
\renewcommand{\arraystretch}{1} 
\resizebox{0.6\linewidth}{!}{
\begin{tabular}{@{}lrrr@{}}
\toprule
\textbf{Benchmark} & \textbf{Train} & \textbf{Val.} & \textbf{Test} \\
\midrule
\cradd{MSRVTT-9k} \cite{msrvtt-1ka} & 9,000 & -- & 1,000 \\
\cradd{MSRVTT-7k} \cite{msrvtt} & 6,513 & 497 & 2,990 \\
VATEX \cite{vatex} & 25,991 & 1,500 & 1,500 \\
Charades \cite{charades} & 7,985 & -- & 1,863 \\
LSMDC \cite{lsmdc} & 101,079 & 7,408 & 1,000 \\
\bottomrule
\end{tabular}}
\label{tab:dataset}
\end{table}

\textbf{Evaluation criteria}. We report standard rank-based retrieval metrics, \ie Recall at top $k$ ($\{1, 5, 10\}$), and SumR (R1+R5+R10) as an overall score.

\textbf{Implementation details}. We initialize the $\mbox{CLIP}_{vis}$ and $\mbox{CLIP}_{txt}$ with OpenAI-released CLIP \cite{clip} (ViT-B/32) weights, and AST with pretrained weights on AudioSet and ImageNet. The two $\mbox{CLIP}_{txt}$ encoders in speech branch and query branch share parameters during training. To reduce GPU memory overhead, the parameters of the audio encoder are kept frozen during training. The model is trained for up to 5 epochs using an Adam optimizer \cite{adam} and a cosine learning rate decay schedule \cite{sgdr} on 8 NVIDIA RTX 3090 GPUs. To prevent catastrophic forgetting, we fine-tune the CLIP backbone with a low learning rate of 1e-7, while other trainable modules are trained with a rate of 1e-4. For MSRVTT, VATEX, and LSMDC datasets, we set the max words number, max frame number and batch size to 32, 12, 128. Since videos and captions in Charades are longer and more complex, we set the max words number, max frame number and batch size to 64, 32, and 64. During testing, for datasets with validation splits (\cradd{MSRVTT-7k}, VATEX, LSMDC), we select the checkpoint with the best R1 on the validation set. As for \cradd{MSRVTT-9k} and Charades which have no validation set available, we follow \cite{clip4clip,teachclip,eclipse,tefal,avigate} and report the peak test performance.

\subsection{Comparison with SOTA} \label{ssec:eval-stoa}

\begin{table*}[!htbp]
\centering
\caption{\textbf{Comparing with visual and audiovisual models for text-to-video retrieval}. 
Per baseline, we cite numbers from its original paper where applicable. The proposed \modelname model outperforms visual and audiovisual baselines on all datasets.  
}
\label{tab:baselines}
\setlength{\tabcolsep}{3pt} 
\renewcommand{\arraystretch}{1.1} 
\resizebox{\linewidth}{!}{
\begin{tabular}{@{}lrrrr|rrrr|rrrr|rrrr|rrrrr@{}}
\toprule

\multirow{2}{*}{\textbf{Model}} & \multicolumn{4}{c}{\textbf{\cradd{MSRVTT-9k}}} &\multicolumn{4}{c}{\textbf{\cradd{MSRVTT-7k}}} & \multicolumn{4}{c}{\textbf{VATEX}} & \multicolumn{4}{c}{\textbf{Charades}} &
\multicolumn{4}{c}{\textbf{LSMDC}} & \multirow{2}{*}{\textbf{mR1}}\\ 

\cmidrule(r){2-5} \cmidrule(r){6-9} \cmidrule(r){10-13} \cmidrule(r){14-17} \cmidrule(r){18-21}
& R1 & R5 & R10 & SumR & R1 & R5 & R10 & SumR & R1 & R5 & R10 & SumR & R1 & R5 & R10 & SumR & R1 & R5 & R10 & SumR & \\ 
\midrule
\rowcolor{gray!20}
\multicolumn{22}{@{}l}{\textit{\textbf{Vision-only:}}}\\
CLIP4Clip, Neucom22\cite{clip4clip}    & 44.5 & 71.4 & 81.6 & 197.5 & 29.4 & 54.9 & 65.8 & 150.1 & 61.6 & 91.1 & 95.8 & 248.5 & 17.7 & 39.5 & 50.4 & 107.6 & 22.6 & 41.0 & 49.1 & 112.7 & 35.1 \\
X-Pool, CVPR22\cite{xpool}      & 46.9 & 72.8 & 82.2 & 201.9 &  -   &  -   &  -   &  -   &  -   &  -   &  -   &  -   &  -   &  -   &  -   &  -   & 25.2 & 43.7 & 53.5 & 122.4 & - \\
TS2-Net, ECCV22\cite{ts2net}      & 47.0 & 74.5 & 83.8 & 205.3 & 29.9 & 56.4 & 67.3 & 153.6 & 59.1 & 90.0 & 95.2 & 244.3 & 16.9 & 38.5 & 49.5 & 104.9 & 23.4 & 42.3 & 50.9 & 116.6 & 35.3 \\
X-CLIP, MM22\cite{xclip}      & 46.1 & 73.0 & 83.1 & 202.2 & 31.2 & 57.4 & 68.1 & 156.7 & 62.2 & 90.9 & 95.4 & 248.5 & 18.9 & 41.3 & 53.0 & 113.2 & 23.3 & 43.0 & - & - & 36.3 \\
CLIP-ViP, ICLR23\cite{clipvip}    & 46.5 & 72.1 & 82.5 & 201.1 & 30.2 & 56.0 & 69.1 & 155.3 & 62.1 & 90.2 & 97.0 & 249.3 &  -   &  -   &  -   &  -   & - & - & - & - & - \\
STAN, CVPR23\cite{stan}       & 46.9 & 72.8 & 82.8 & 202.5 &  -   &  -   &  -   &  -   &  -   &  -   &  -   &  -   &  -   &  -   &  -   &  -   & 23.7 & 42.7 & 51.8 & 118.2 & - \\
DiCoSA, IJCAI23\cite{dicosa}       & 47.5 & 74.7 & 83.8 & 206.0 &  -   &  -   &  -   &  -   & - & - & - & - &  -   &  -   &  -   &  -   & 25.4 & 43.6 & 54.0 & 123.0 & - \\
PromptSwitch, ICCV23\cite{promptswitch} & 47.8 & 73.9 & 82.2 & 203.9 &  -   &  -   &  -   &  -   &  -   &  -   &  -   &  -   &  -   &  -   &  -   &  -   & 23.1 & 41.7 & 50.5 & 115.3 & - \\
UATVR, ICCV23\cite{uatvr}       & 47.5 & 73.9 & 83.5 & 204.9 &  -   &  -   &  -   &  -   & 61.3 & 91.0 & 95.6 & 247.9 &  -   &  -   &  -   &  -   &  -   &  -   &  -   &  -  & - \\
UCoFiA, ICCV23\cite{ucofia}      & 48.2 & 73.3 & 82.3 & 203.8 & 29.9 & 56.0 & 66.7 & 152.6 & 61.1 & 90.5 &  -   &   -   & 17.6 & 38.6 & 48.6 & 104.8 & 22.5 & 41.6 & 51.4 & 115.5 & 35.9 \\
ProST, ICCV23\cite{prost}        & 48.2 & 74.6 & 83.4 & 206.2 & 30.4 & 56.7 & 67.7 & 154.8 & 60.6 & 90.5 & 95.4 & 246.5 & 17.8 & 37.0 & 48.3 & 103.1 & 24.1 & 42.5 & 51.6 & 118.2 & 36.2 \\
DGL, AAAI24\cite{dgl}        & 45.8 & 69.3 & 79.4 & 194.5 & - & - & - & - & 56.2 & 87.1 & 93.5 & 236.8 &  -   &  -   &  -   &  -  &  21.4 & 39.4 & 48.4 & 109.2 & - \\
EERCF, AAAI24\cite{eercf}        & 47.8 & 74.1 & 84.1 & 206.0 & 31.5 & 57.4 & 67.6 & 156.5 & 62.6 & 91.5 & 95.8 & 249.9 &  -   &  -   &  -   &  -   &  -   &  -   &  -   &  -  & -  \\
TeachCLIP, CVPR24\cite{teachclip}    & 46.8 & 74.3 & 82.6 & 203.7 & 30.9 & 57.1 & 68.0 & 156.0 & 63.6 & 91.9 & 96.1 & 251.6 & 19.2 & 40.1 & 51.2 & 110.5 &  -   &  -   &  -   &  -  & - \\
TempMe, ICLR25\cite{tempme}      & 46.1 & 71.8 & 80.7 & 198.6 &  -   &  -   &  -   &  -   &  -   &  -   &  -   &  -   &  -   &  -   &  -   &  -   & 23.5 & 41.7 & 51.8 & 117.0 & - \\
DiscoVLA, CVPR25\cite{discovla}    & 47.0 & 73.0 & 82.8 & 202.8 &  -   &  -   &  -   &  -   &  -   &  -   &  -   &  -   &  -   &  -   &  -   &  -   & 23.6 & 42.0 & 52.3 & 117.9 & - \\
PIG, ICCV25\cite{pig}         & 48.6 & 72.8 & 81.6 & 203.0 & 31.8 & 57.3 & 68.0 & 157.1 & 64.0 & 91.5 & 96.6 & 252.1 &  -   &  -   &  -   &  -   &  -   &  -   &  -   &  -  & - \\

\midrule
\rowcolor{gray!20}
\multicolumn{22}{@{}l}{\textit{\textbf{Audiovisual:}}}\\
EclipSE, ECCV22\cite{eclipse}      & 44.9 & 71.3 & 81.6 & 197.8 & 30.2 & 55.9 & 66.6 & 152.7 & 57.8 & 88.4 & 94.3 & 240.5 & 15.7 & - & - & - & 22.2 & 43.8 & 52.9 & 118.9 & 34.2 \\

TEFAL, ICCV23\cite{tefal}        & 49.4 & 75.9 & 83.9 & 209.2 & 32.6 & 59.1 & 69.3 & 161.0 & 61.0 & 90.4 & 95.3 & 246.7 & 18.5 & 37.3 & 48.6 & 104.4 & 24.7 & 45.1 & 53.7 & 123.5 &  37.2 \\
AVIGATE, CVPR25\cite{avigate}      & 50.2 & 74.3 & 83.2 & 207.7 & 32.7 & 59.8 & 70.2 & 162.7 & 63.1 & 90.7 & 95.5 & 249.3 & 18.8 & 40.0 & 51.8 & 110.6 & 24.6 & 46.0 & 55.1 & 125.7 & 37.9 \\
AVIGATE-h \cite{avigate}      & 46.1 & 73.4 & 82.4 & 201.9 & 32.0 & 58.6 & 69.1 & 159.7 & 60.8 & 90.1 & 95.4 & 246.3 & 18.6 & 39.5 & 51.5 & 109.6 & 22.8 & 43.7 & 53.5 & 120.0 & 36.1 \\
AVIGATE+     & 50.9 & 75.8 & 85.2 & 211.9 & 33.0 & 60.3 & 70.8 & 164.1 & 64.8 & 92.3 & 96.6 & 253.7 & 19.9 & 42.3 & 54.4 & 116.6 & 25.5 & 45.7 & 55.1 & 126.3 & 38.8 \\
\rowcolor{green!20}
SAVE         & \textbf{51.3} & \textbf{78.0} & \textbf{86.9} & \textbf{216.2} & \textbf{33.5} & \textbf{60.9} & \textbf{71.4} & \textbf{165.8} & \textbf{66.1} & \textbf{92.6} & \textbf{96.8} & \textbf{255.5} & \textbf{20.8} & \textbf{44.7} & \textbf{55.9} & \textbf{121.4} & \textbf{26.1} & \textbf{46.4} & \textbf{55.8} & \textbf{128.3} & \textbf{39.6} \\
SAVE-h   & 49.2 & 77.0 & 85.1 & 211.3 & 32.9 & 60.0 & 70.4 & 163.3 & 64.5 & 91.6 & 96.0 & 252.1 & 19.3 & 41.2 & 53.5 & 114.0 & 24.4 & 43.8 & 53.4 & 121.6 & 38.1 \\

\bottomrule
\end{tabular}
}
\end{table*}

\textbf{Baselines}. Both visual and audiovisual methods are compared. For the purpose of fair comparison and reproducibility of research, we only include methods that do not rely on additional training data and are open-sourced, as follows:\\
$\bullet$ \emph{Vision-only}: CLIP4Clip \cite{clip4clip}, X-Pool \cite{xpool}, TS2-Net \cite{ts2net}, X-CLIP \cite{xclip}, CLIP-ViP \cite{clipvip}, STAN \cite{stan}, DiCoSA \cite{dicosa}, PromptSwitch \cite{promptswitch}, UATVR \cite{uatvr}, UCoFiA \cite{ucofia}, ProST \cite{prost}, DGL \cite{dgl}, EERCF \cite{eercf}, TeachCLIP \cite{teachclip}, TempMe \cite{tempme}, DiscoVLA \cite{discovla}, and PIG \cite{pig}.\\
$\bullet$ \emph{Audiovisual}: EclipSE \cite{eclipse}, TEFAL \cite{tefal}, and our backbone baseline AVIGATE \cite{avigate}. We further consider two AVIGATE variants: \mbox{AVIGATE-h}, which only uses holistic similarity $s_g$ to generate the video-text similarity matrix, and \mbox{AVIGATE+}, which equips AVIGATE with our soft-ALBEF alignment strategy. We also provide a holistic variant of our \modelname{}, termed \mbox{SAVE-h}, which employs LAFF \cite{laff} to aggregate $\{\tilde{v}_i\}$ into a holistic video representation and also only computes holistic similarity.


\textbf{Performance comparison}. As shown in \cref{tab:baselines}, AVIGATE is the state-of-the-art baseline among audiovisual methods. Our \modelname{} achieves new SOTA performance on text-to-video retrieval across all datasets. Specifically, \modelname{} outperforms AVIGATE by 8.5 on \cradd{MSRVTT-9k}, 3.1 on \cradd{MSRVTT-7k}, 6.2 on VATEX, 10.8 on Charades, and 2.6 on LSMDC in SumR, and 1.7 in mR1 (mean R1 across all datasets). The consistent improvements across different datasets validate the effectiveness of our method. Notably, despite only 13.5\% of videos in the Charades dataset containing ASR transcripts, our \modelname{} still achieves a substantial 10.8 SumR improvement over AVIGATE. We attribute it to our soft-ALBEF alignment strategy that better leverages sound information in the audio modality. \cradd{Comparing \modelname and AVIGATE+ reveals the gain of the speech branch.} Our SAVE-h variant also yields comparable performance using only holistic video representations, further demonstrating the robustness of our design. 


\textbf{Efficiency Analysis}. Let $n_\mathcal{T}$, $n_\mathcal{V}$, $n_\mathcal{A}$, and $n_\mathcal{S}$ be the number of text queries, videos, audios, and speech transcripts, respectively, where $n_\mathcal{V} \ge n_\mathcal{A} \ge n_\mathcal{S}$. As shown in \cref{tab:efficiency}, 
the computational complexity reflects the theoretical scaling behavior \wrt the number of multimodal samples, while the inference time measures the actual online latency for a single text query embedding extraction and similarity computation between the query embedding and all pre-extracted video representations. Methods such as X-Pool and TEFAL rely on query-dependent interactions, resulting in quadratic computational complexity. In contrast, \modelname{} maintains linear complexity, ensuring scalability to large video collections. Despite the additional speech branch, \modelname{} achieves the same inference latency as AVIGATE, as all video features can be extracted offline and thus do not contribute to the online inference cost.

\begin{table}[ht]
\caption{\textbf{Efficiency analysis}.
Inference time includes both query feature extraction and video-text similarity computation. Test set: \cradd{MSRVTT-9k}. GPU: NVIDIA RTX 3090.}
\centering
\setlength{\tabcolsep}{5pt} 
\renewcommand{\arraystretch}{1} 
\resizebox{\linewidth}{!}{
\label{tab:efficiency}
\begin{tabular}{@{}llrr@{}}
\toprule
\textbf{Methods} & \textbf{Computational complexity} & \textbf{Inference time (ms)}$\downarrow$ & \textbf{SumR}$\uparrow$ \\
\midrule
\rowcolor{gray!20}
\multicolumn{4}{@{}l}{\textit{\textbf{Visual:}}} \\
CLIP4Clip \cite{clip4clip} & $\mathcal{O}(n_\mathcal{V} + n_\mathcal{T})$ & 9.76 & 197.5  \\
X-Pool \cite{xpool} & $\mathcal{O}(n_\mathcal{V}n_\mathcal{T})$ & 66.31 & 201.9 \\
PIG \cite{pig} & $\mathcal{O}(n_\mathcal{V} + n_\mathcal{T})$  & 9.76 & 203.0 \\
\midrule
\rowcolor{gray!20}
\multicolumn{4}{@{}l}{\textit{\textbf{Audiovisual:}}} \\
TEFAL \cite{tefal} & $\mathcal{O}(n_\mathcal{A}n_\mathcal{T} + n_\mathcal{V}n_\mathcal{T})$ & 140.57 & 209.2 \\
AVIGATE \cite{avigate} & $\mathcal{O}(n_\mathcal{A} + n_\mathcal{V} + n_\mathcal{T})$ & 9.90 & 207.7 \\
AVIGATE-h \cite{avigate} & $\mathcal{O}(n_\mathcal{A} + n_\mathcal{V} + n_\mathcal{T})$ & 9.76 & 201.9 \\

\rowcolor{green!20}
SAVE & $\mathcal{O}(n_\mathcal{S} + n_\mathcal{A} + n_\mathcal{V} + n_\mathcal{T})$ & 9.90 & \textbf{216.2} \\
SAVE-h & $\mathcal{O}(n_\mathcal{S} + n_\mathcal{A} + n_\mathcal{V} + n_\mathcal{T})$ & 9.76 & 211.3 \\
\bottomrule
\end{tabular}}
\end{table}

\begin{figure*}[!htbp]
  \centering
  \begin{subfigure}[b]{0.48\textwidth}
    \centering
    \includegraphics[width=\linewidth]{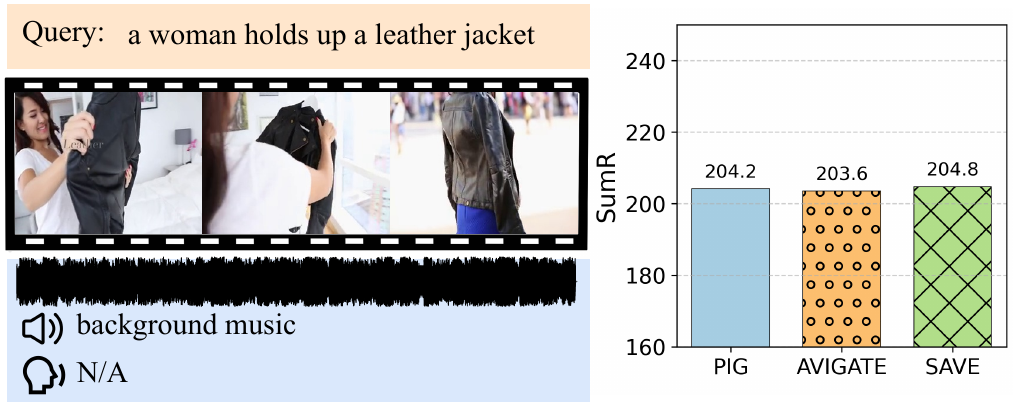}
    \caption{Visual-based (499 cases)}
    \label{fig:group1}
  \end{subfigure}
  \hspace{4mm}
  \begin{subfigure}[b]{0.48\textwidth}
    \centering
    \includegraphics[width=\linewidth]{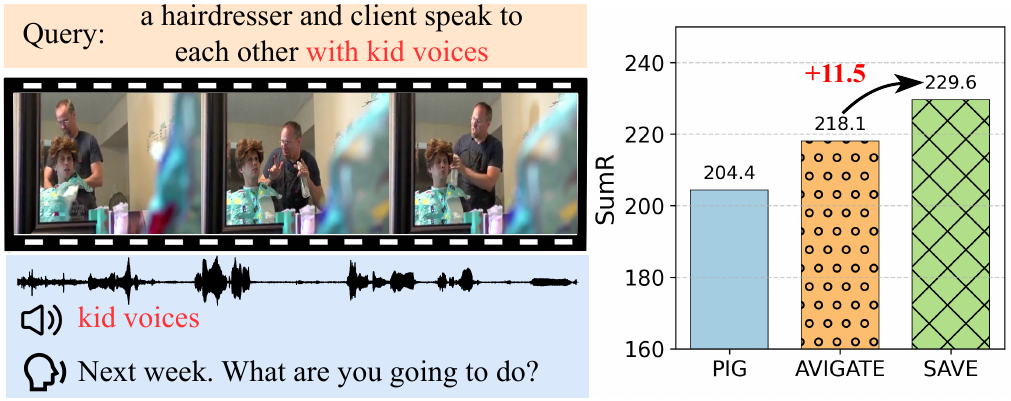}
    \caption{Sound-related (226 cases)}
    \label{fig:group2}
  \end{subfigure}
  
  \vspace{5.5mm}
  \begin{subfigure}[b]{0.48\textwidth}
    \centering
    \includegraphics[width=\linewidth]{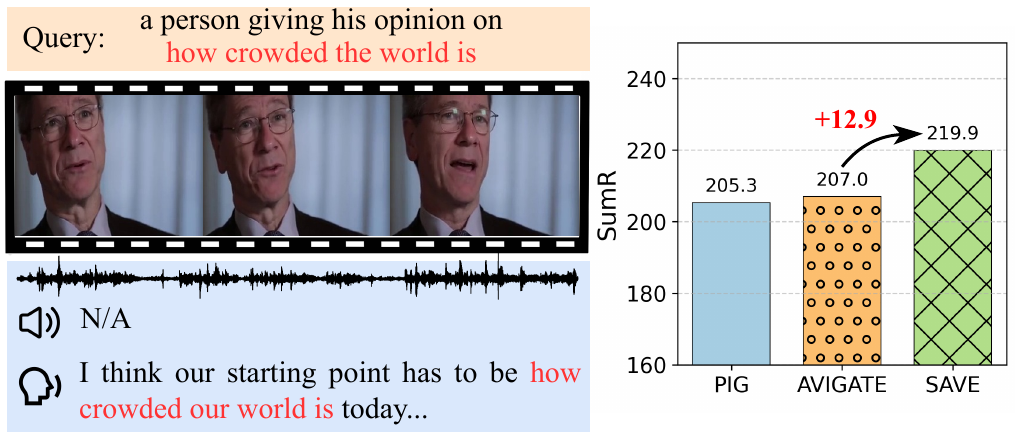}
    \caption{Speech-related (171 cases)}
    \label{fig:group3}
  \end{subfigure}
  \hspace{4mm}
  \begin{subfigure}[b]{0.48\textwidth}
    \centering
    \includegraphics[width=\linewidth]{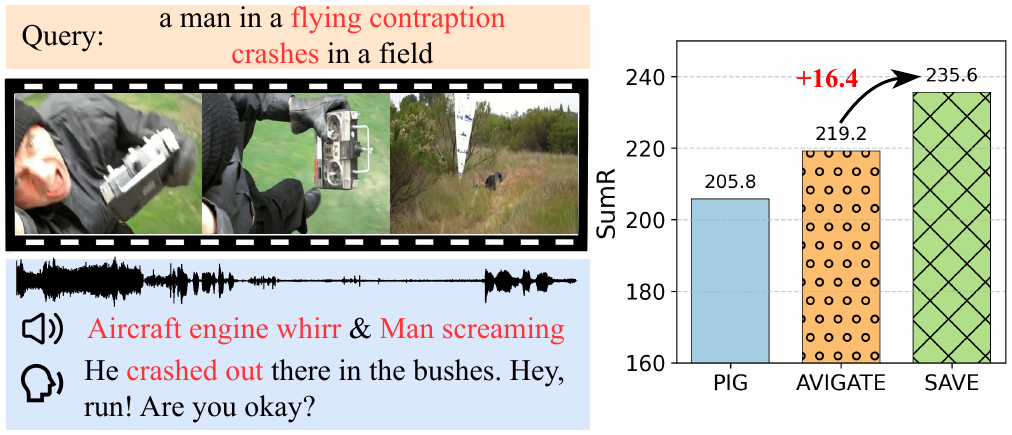}
    \caption{Sound-Speech-related (104 cases)}
    \label{fig:group4}
  \end{subfigure}
  \caption{\textbf{Comparison per group}. 
  Our \modelname consistently outperforms PIG (best visual model) and  AVIGATE (best audiovisual model) across all groups, 
  with the largest gain obtained in the Sound-Speech-related group.
  }
  \label{fig:group}
\end{figure*}

\subsection{Understanding \modelname{}}
\label{ssec:ablation}

\modelname{} achieves superior performance through a dedicated speech branch and soft-ALBEF early vision-audio alignment. 
To better understand the two designs,
we provide 
a series of ablation studies as follows. 

\textbf{Comparison per group}. We begin with a fine-grained group comparison. To that end, we manually categorize the test samples of \cradd{MSRVTT-9k} into four mutually exclusive groups, depending on whether a specific query is related to sound and/or speech: (a) visual-based, (b) sound-related, (c) speech-related, and (d) sound-speech-related. As shown in \cref{fig:group}, \modelname consistently outperforms PIG and AVIGATE across all groups. For the sound-related group, \modelname{} achieves an 11.5 improvement in SumR over AVIGATE, primarily stemming from our proposed soft-ALBEF alignment strategy. For the speech-related group, the substantial gain of 12.9 over AVIGATE arises from the newly introduced speech branch that explicitly models semantic information in dialogue. Notably, in the sound-speech-related group where both acoustic and semantic cues jointly contribute, \modelname{} achieves the largest improvement of 16.4 over AVIGATE.

\textbf{Speech and sound branches both benefit video-text retrieval}. As shown in \cref{tab:ablations}, removing the speech branch (Setup\#1) and sound branch (Setup\#4) results in performance drops of 4.3 and 8.7 points in SumR on \cradd{MSRVTT-9k}, respectively. These results demonstrate that both branches contribute meaningfully to the model's retrieval performance. While the degradation caused by removing the sound branch appears larger, this is mainly because \cradd{MSRVTT-9k} contains notably more sound-related queries than speech-related ones (see \cref{fig:group}).

\begin{table}[!hbtp]
\caption{\textbf{Ablation studies of \modelname{}}. }
\centering
\setlength{\tabcolsep}{2.5pt} 
\renewcommand{\arraystretch}{1.1} 
\resizebox{0.98\linewidth}{!}{
\begin{tabular}{@{}clllcll@{}}
\toprule
\multirow{2}{*}{\textbf{\#}} & \multirow{2}{*}{\textbf{Setup}} & \multicolumn{2}{c}{\textbf{\cradd{MSRVTT-9k}}} & & \multicolumn{2}{c}{\textbf{\cradd{MSRVTT-7k}}} \\
\cmidrule{3-4} \cmidrule{6-7}
 & & \multicolumn{1}{c}{R1} & \multicolumn{1}{l}{SumR} & & \multicolumn{1}{c}{R1} & \multicolumn{1}{l}{SumR} \\
\midrule
\rowcolor{green!20}
0 & Full & \textbf{51.3} & \textbf{216.2} & & \textbf{33.5} & \textbf{165.8} \\
\midrule
\rowcolor{gray!20}
\multicolumn{7}{@{}l}{\textbf{\textit{Speech branch:}}} \\
1 & \textit{w/o} Speech Branch & 50.9 & 211.9(\drop{4.3}$\downarrow$) & & 33.0 & 164.1(\drop{1.7}$\downarrow$) \\
2 & ASR: Whisper $\rightarrow$ SenseVoice & 51.2 & 215.4(\drop{0.8}$\downarrow$) & & 33.5 & 165.4(\drop{0.4}$\downarrow$) \\
3 & $\mbox{CLIP}_{txt}$: \textit{w/o} parameter sharing & 51.0 & 214.8(\drop{1.4}$\downarrow$) & & 33.4 & 165.0(\drop{0.8}$\downarrow$) \\
\midrule
\rowcolor{gray!20}
\multicolumn{7}{@{}l}{\textbf{\textit{Sound branch:}}} \\
4 & \textit{w/o} Sound Branch & 48.5 & 207.5(\drop{8.7}$\downarrow$) & & 32.0 & 159.3(\drop{6.5}$\downarrow$) \\
5 & \textit{w/o} Soft-ALBEF & 49.3 & 213.8(\drop{2.4}$\downarrow$) & & 33.3 & 164.8(\drop{1.0}$\downarrow$) \\
6 & Soft-ALBEF $\rightarrow$ ALBEF & 49.2 & 211.2(\drop{5.0}$\downarrow$) & & 33.2 & 164.6(\drop{1.2}$\downarrow$) \\
7 & \cradd{Pearson $\rightarrow$ MSE} & 50.8 & 215.0(\drop{1.2}$\downarrow$) & & 33.4 & 165.3(\drop{0.5}$\downarrow$) \\
8 & \cradd{Pearson $\rightarrow$ Huber} & 50.8 & 215.7(\drop{0.5}$\downarrow$) & & 33.5 & 165.5(\drop{0.3}$\downarrow$) \\
9 & \cradd{ImageBind $\rightarrow$ AudioCLIP} & 51.1 & 215.8(\drop{0.4}$\downarrow$) & & 33.6 & 165.6(\drop{0.2}$\downarrow$) \\
\midrule
\rowcolor{gray!20}
\multicolumn{7}{@{}l}{\textbf{\textit{Fusion:}}} \\
10 & \cradd{Late-fusion} & 47.8 & 206.5(\drop{9.7}$\downarrow$) & & 31.4 & 155.6(\drop{10.2}$\downarrow$) \\
11 & \cradd{Learnable weights} & 51.2 & 215.5(\drop{0.7}$\downarrow$) & & 33.4 & 165.3(\drop{0.5}$\downarrow$) \\
\bottomrule
\end{tabular}
}
\label{tab:ablations}
\end{table}

\textbf{Choice of ASR models}. We replace Whisper with an alternative ASR model, SenseVoice \cite{sensevoice}, to evaluate the impact of ASR backbone choice. Setup\#2 shows that Whisper is the optimal choice for our framework.

\textbf{Using the same text encoder for query and speech embedding}?
Since the text encoder in the speech branch and that in the query encoder are both instantiated by CLIP,  a natural question is whether they shall share the same weights. 
The results of Setup\#3 indicates that sharing is preferred. 
Our interpretation is that parameter sharing helps reduce the gap between the speech and query modalities, thereby facilitating more effective cross-modal retrieval.

\textbf{The role of early vision-audio alignment}. Directly applying ABLEF for vision-audio alignment even leads to performance degradation compared with no alignment (Setup \#5 \emph{vs} \#6). This indicates that making video-audio pairs for early alignment is non-trivial. By contrast, our soft-ABLEF strategy offers a feasible solution.

\begin{figure}[!htbp]
    \centering
    \includegraphics[width=0.7\columnwidth]{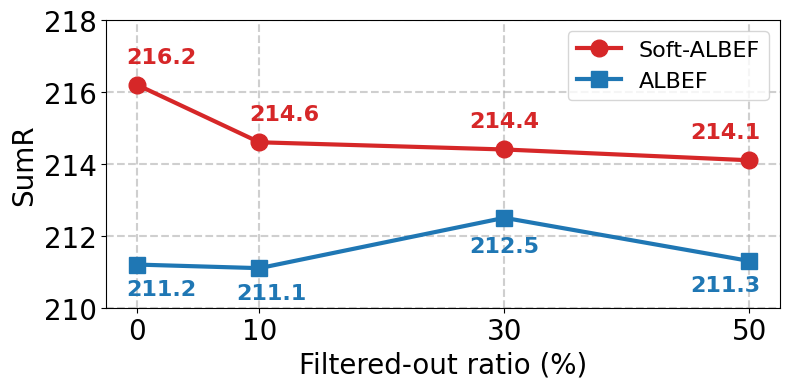}
    \caption{Soft-ALBEF \emph{vs} ALBEF with \emph{low-score} data filtered out.}
    \label{fig:filter}
\end{figure}

\cradd{\textbf{Comparison with data filtering.} A simpler alternative to soft-ALBEF is to filter out \emph{low-ImageBind-score} video-audio pairs before applying ALBEF. As shown in \cref{fig:filter}, soft-ALBEF consistently outperforms ALBEF across all filtered-out ratios, demonstrating its effectiveness.}

\cradd{\textbf{Which loss for soft-ALBEF}? As shown in \cref{tab:ablations}, the results of Setup\#7 and Setup\#8 indicate that Pearson distance loss is a better choice than MSE and Huber loss.}

\cradd{\textbf{Choice of teacher model}. Setup\#9 replaces ImageBind with a weaker teacher AudioCLIP \cite{audioclip}, and still yields a clear gain over AVIGATE, confirming that the benefit comes from soft-labeling rather than the specific teacher.}

\cradd{\textbf{Late fusion \vs early fusion.} Setup\#10 implements a late-fusion baseline, where the speech-query similarity score is computed separately and averaged with the video-audio-query similarity score. The performance drop demonstrates the superiority of our early fusion strategy.}

\cradd{\textbf{Learnable weights for obtaining $\{\tilde{v}_i\}$.} Setup\#11 replaces the parameter-free fusion $\{v_i\} + (\{\hat{a}_i\} + \{\hat{s}_i\}) / 2$ with a parameterized version $\alpha\{v_i\} + \beta\{\hat{a}_i\} + (1-\alpha-\beta)\{\hat{s}_i\}$ using learnable weights $\alpha$ and $\beta$. The performance does not improve, validating our motivation discussed in \cref{ssec:speech}.}

\textbf{Using ImageBind directly for VTR}? We employ ImageBind to generate soft labels for soft-ALBEF. 
A natural question arises as \textit{Is ImageBind itself suitable for VTR}? We evaluate ImageBind under three different modality configurations for constructing video representations: \\
$\bullet$ ImageBind-v: Video representation is derived from the visual modality by averaging frame-level features. \\
$\bullet$ ImageBind-va: Video representation is obtained by combining visual and audio features. \\
$\bullet$ ImageBind-vas: Incorporates video, audio, and speech features jointly for video representation. 

Note that ImageBind uses ViT-H/14, which is too large to be fully finetuned for VTR.
To investigate its applicability within our GPU budget (8$\times$ RTX 3090 GPUs). We evaluate all three variants in a zero-shot manner. Additionally, for ImageBind-vas, we further explore a partial finetuning strategy, where the vision and audio encoders are frozen except for their linear projection layers, while the text encoder, computationally lightweight, is fully finetuned.

As shown in \cref{tab:imagebind}, ImageBind performs worse than our SAVE (ViT-B/32) under both settings, suggesting that ImageBind is not directly suitable as a backbone for video-text retrieval, especially under limited computational resources. However, this conclusion does not conflict with its specialized use in our soft-ALBEF strategy. For soft-ALBEF, we do not use ImageBind as the primary retrieval backbone. Instead, we leverage its powerful vision-audio alignment capabilities only to compute a high-quality similarity matrix.

\begin{table}[!hbtp]
\caption{\textbf{Comparison with ImageBind}. Dataset: \cradd{MSRVTT-9k}.}
\centering
\setlength{\tabcolsep}{8pt} 
\renewcommand{\arraystretch}{1} 
\resizebox{0.7\linewidth}{!}{
\begin{tabular}{@{}lrrrr}
\toprule
\textbf{Methods} & R1 & R5 & R10 & SumR \\
\midrule
\rowcolor{gray!20}
\multicolumn{5}{@{}l}{\textbf{\textit{Zero-shot:}}} \\
ImageBind-v & 39.1 & 62.8 & 73.1 & 175.0 \\
ImageBind-va & 39.4 & 63.2 & 73.7 & 176.3 \\
ImageBind-vas & 40.6 & 62.6 & 74.1 & 177.3 \\
\midrule
\rowcolor{gray!20}
\multicolumn{5}{@{}l}{\textbf{\textit{Finetuned:}}} \\
ImageBind-vas & 47.2 & 72.9 & 82.0 & 202.1 \\
\rowcolor{green!20}
SAVE & \textbf{51.3} & \textbf{78.0} & \textbf{86.9} & \textbf{216.2} \\
\bottomrule
\end{tabular}
}
\label{tab:imagebind}
\end{table}

\section{Conclusions}
\label{sec:conclusion}

We propose \modelname{}, a \underline{s}peech-\underline{a}ware \underline{v}ideo r\underline{e}presentation for audio-enhanced video-text retrieval (VTR). Our work identifies two key limitations in current methods for audio-enhanced VTR, namely the under-exploitation of speech semantics and the absence of early video-audio alignment, and accordingly proposes two targeted, effective solutions. Extensive experiments on multiple benchmarks allow us to conclude as follows. The dedicated speech branch effectively captures semantic information from dialog and provides complementary benefits to the sound branch. The soft-ALBEF strategy offers an effective solution for early vision-audio alignment when direct pairing is noisy or unreliable. This study opens a promising direction for the integration of speech with other types of audio cues for multimodal VTR. 

\textbf{Limitation of this study}. Our experiments are conducted on short video clips with brief ASR transcripts that can be directly handled by the CLIP text encoder. In more complex scenarios, such as e-commerce livestream understanding, ASR transcripts are typically much longer and noisier. How to efficiently extract key information from such lengthy and noisy speech to obtain robust speech-aware video representation warrants further research.

\medskip
\textbf{Acknowledgments}. This research was supported by NSFC (No. 62576348) and Beijing Natural Science Foundation (No. L254039).

{
    \small
    \bibliographystyle{ieeenat_fullname}
    \bibliography{main}
}

\clearpage
\setcounter{page}{1}
\setcounter{section}{0}
\setcounter{table}{0}
\setcounter{figure}{0}
\renewcommand{\thesection}{S\arabic{section}}
\renewcommand{\thefigure}{S\arabic{figure}}
\renewcommand{\thetable}{S\arabic{table}}


\maketitlesupplementary

In this supplementary material, we provide additional experiments not included in the main paper, including:\\
$\bullet$ t-SNE visualization of more audio encoders (\cref{sup_sec:tsne}) \\
$\bullet$ Using Whisper as the audio encoder (\cref{sup_sec:whisper}) \\
$\bullet$ Video-to-Text retrieval results (\cref{sup_sec:v2t})\\
$\bullet$ Using a larger backbone (\cref{sup_sec:vit_b_16})\\
$\bullet$ Qualitative results (\cref{sup_sec:qualitative})

\section{Visualization of More Audio Encoders}
\label{sup_sec:tsne}

\begin{figure}[!htbp]
  \centering
  \begin{subfigure}[b]{\linewidth}
    \centering
    \includegraphics[width=\linewidth]{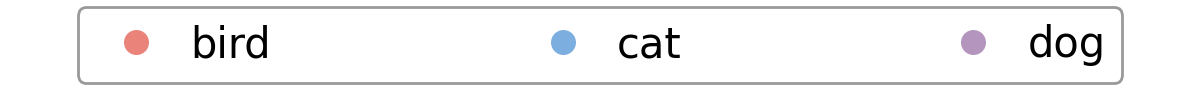}
  \end{subfigure}
  
  \begin{subfigure}[b]{\linewidth}
    \centering
    \includegraphics[width=\linewidth]{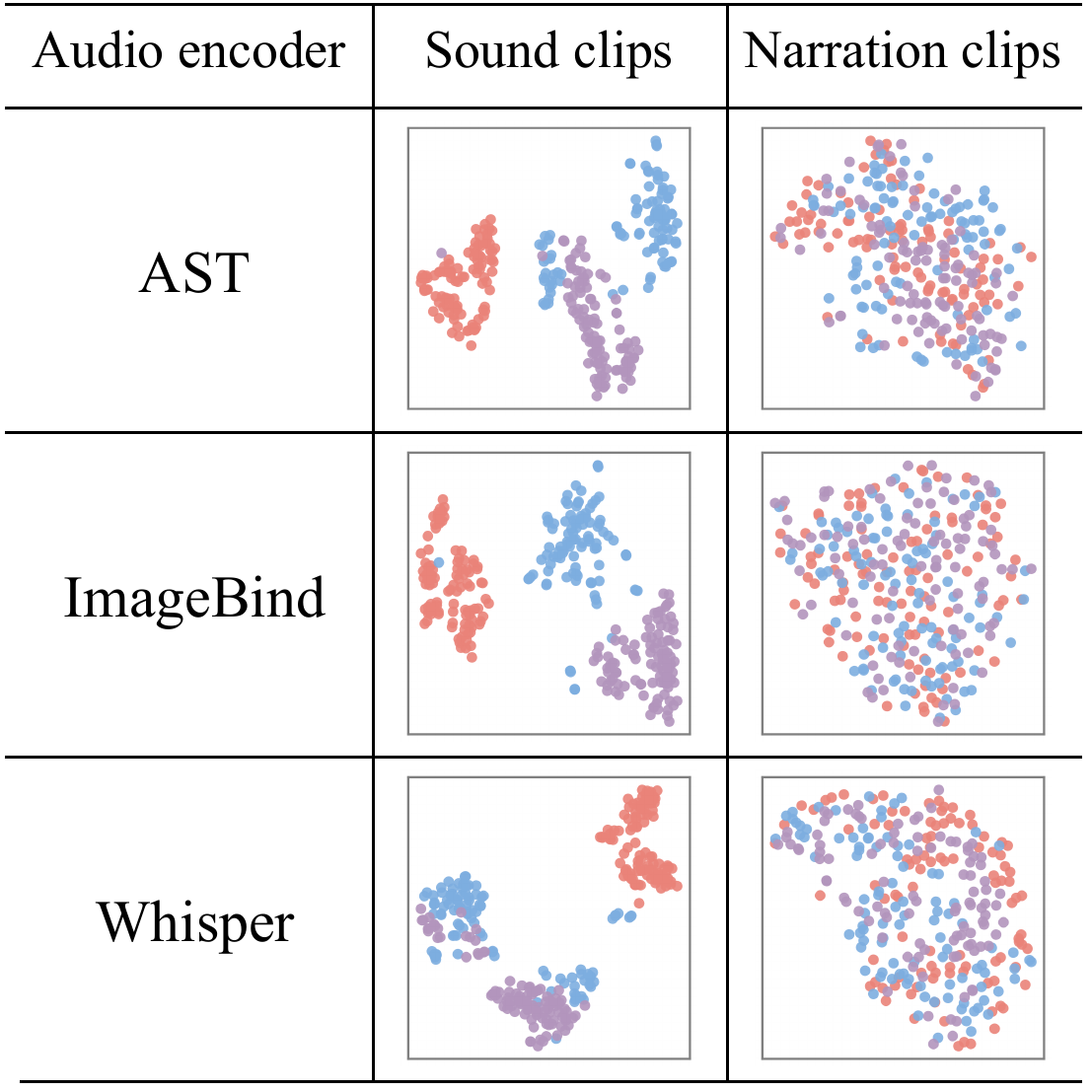}
  \end{subfigure}
  \caption{\textbf{t-SNE visualizations of feature spaces across three audio encoders}. All three audio encoders struggle to capture speech semantics.}
  \label{fig:audioencoder}
\end{figure}

To verify that the lack of speech semantic understanding observed in AST \cite{ast} is a general issue, we extend our analysis to ImageBind \cite{imagebind} and Whisper \cite{whisper}. Following the same experimental setup as Fig. \textcolor[rgb]{.286,.481,.722}{1a} in the main paper, we visualize the audio embeddings for \textit{sound clips} and \textit{narration clips} across three animal classes using t-SNE. As shown in \cref{fig:audioencoder}, while all three models produce distinct clusters for \textit{sound clips}, the embeddings for \textit{narration clips} remain heavily entangled across all encoders. Notably, even Whisper, despite its robust ASR capabilities, fails to disentangle speech semantics in its audio feature space. This confirms that current audio backbones predominantly capture acoustic characteristics rather than high-level speech semantics.


\section{Using Whisper as the Audio Encoder}
\label{sup_sec:whisper}

\cref{tab:whisper} presents an ablation study examining whether a dedicated speech branch remains necessary when using Whisper, which already possesses ASR capability, as the audio encoder. We introduce SAVE-2, which uses Whisper as the audio encoder, and SAVE-3, which further removes the speech branch. Comparing SAVE-2 and SAVE-3, we observe that removing the speech branch consistently degrades performance on both \cradd{MSRVTT-9k} and \cradd{MSRVTT-7k}, indicating that explicit speech modeling remains beneficial even when the audio encoder itself is ASR-capable.


\begin{table}[!hbtp]
\caption{\textbf{Ablation study of using Whisper as the audio encoder}. Backbone: CLIP (ViT-B/32).}
\centering
\setlength{\tabcolsep}{6pt} 
\renewcommand{\arraystretch}{1.1} 
\resizebox{\linewidth}{!}{
\begin{tabular}{@{}lcccrrrr@{}}
\toprule
\multirow{2}{*}{\textbf{Setup}} & \multirow{2}{*}{\textbf{Audio}} & \multirow{2}{*}{\textbf{Speech}} & \multirow{2}{*}{\textbf{Soft-ALBEF}} & \multicolumn{2}{c}{\textbf{\cradd{MSRVTT-9k}}} & \multicolumn{2}{c}{\textbf{\cradd{MSRVTT-7k}}} \\
\cmidrule(r){5-6} \cmidrule(r){7-8}
& & & & R1 & SumR & R1 & SumR \\
\midrule
SAVE & AST & Whisper & \cmark & \textbf{51.3} & \textbf{216.2} & 33.5 & 165.8 \\
SAVE-2 & Whisper & Whisper & \cmark & 50.9 & 215.8 & \textbf{33.8} & \textbf{166.8} \\
SAVE-3 & Whisper & \xmark & \cmark & 50.2 & 213.0 & 32.9 & 164.0 \\
\bottomrule
\end{tabular}
}
\label{tab:whisper}
\end{table}

\section{Video-to-Text Retrieval Results}
\label{sup_sec:v2t}

For a more comprehensive evaluation, we also report video-to-text retrieval performance on \cradd{MSRVTT-9k}, see \cref{tab:v2t}. Same as text-to-video retrieval task (Tab. \textcolor[rgb]{.286,.481,.722}{3}), \modelname{} achieves the highest SumR among all compared methods.

\begin{table}[!hbtp]
\caption{\textbf{The video-to-text retrieval performance of different methods}. Dataset: MSRVTT-9k. Backbone: ViT-B/32.}
\centering
\setlength{\tabcolsep}{4pt} 
\renewcommand{\arraystretch}{1.1} 
\resizebox{0.8\linewidth}{!}{
\begin{tabular}{@{}lrrrr@{}}
\toprule
\textbf{Model} & R1 & R5 & R10 & SumR \\
\midrule
\rowcolor{gray!20}
\multicolumn{5}{@{}l}{\textbf{\textit{Vision-only:}}} \\
CLIP4Clip, Neucom22 \cite{clip4clip} & 43.1 & 70.5 & 81.2 & 194.8 \\
X-Pool, CVPR22 \cite{xpool} & 44.4 & 73.3 & 84.0 & 201.7 \\
TS2-Net, ECCV22 \cite{ts2net} & 45.3 & 74.1 & 83.7 & 203.1 \\
X-CLIP, MM22 \cite{xclip} & 46.8 & 73.3 & 84.0 & 204.1 \\
DiCoSA, IJCAI23 \cite{dicosa} & 46.7 & 75.2 & 84.3 & 206.2 \\
PromptSwitch, ICCV23 \cite{promptswitch} & 46.0 & 74.3 & 84.8 & 205.1 \\
UATVR, ICCV23 \cite{uatvr} & 46.9 & 73.8 & 83.8 & 204.5 \\
UCoFiA, ICCV23 \cite{ucofia} & 47.1 & 74.3 & 83.0 & 204.4 \\
ProST, ICCV23 \cite{prost} & 46.3 & 74.2 & 83.2 & 203.7 \\
DGL, AAAI24 \cite{dgl} & 43.5 & 70.5 & 80.7 & 194.7 \\
EERCF, AAAI24 \cite{eercf} & 44.7 & 74.2 & 83.9 & 202.8 \\
TempMe, ICLR25 \cite{tempme} & 45.6 & 72.4 & 81.2 & 199.2 \\
DiscoVLA, CVPR25 \cite{discovla} & 47.7 & 73.6 & 83.6 & 204.9 \\
\midrule
\rowcolor{gray!20}
\multicolumn{5}{@{}l}{\textbf{\textit{Audiovisual:}}} \\
EclipSE, ECCV22 \cite{eclipse} & 44.7 & 71.3 & 82.8 & 198.8 \\
TEFAL, ICCV23 \cite{tefal} & 47.1 & 75.1 & 84.9 & 207.1 \\
AVIGATE, CVPR25 \cite{avigate} & \textbf{49.7} & 75.3 & 83.7 & 208.7 \\
AVIGATE+ & 49.5 & 75.5 & 85.4 & 210.4 \\
\rowcolor{green!20}
SAVE & 48.9 & \textbf{78.0} & \textbf{86.5} & \textbf{213.4} \\
SAVE-h   & 46.7 & 76.0 & 85.6 & 208.3 \\
\bottomrule
\end{tabular}
}
\label{tab:v2t}
\end{table}

\section{Using a Larger Backbone}
\label{sup_sec:vit_b_16}

To verify if our method's superiority holds with a larger backbone, we repeat the main experiments using CLIP (ViT-B/16). As shown in \cref{tab:vitb16}, our \modelname{} consistently outperforms both vision-only and audiovisual baselines.

\begin{table}[!hbtp]
\caption{\textbf{The retrieval performance of different methods on CLIP ViT-B/16 backbone}. Dataset: MSRVTT-9k.}
\centering
\setlength{\tabcolsep}{4pt} 
\renewcommand{\arraystretch}{1.1} 
\resizebox{\linewidth}{!}{
\begin{tabular}{@{}lrrrr|rrrr@{}}
\toprule

\multirow{2}{*}{\textbf{Model}} & \multicolumn{4}{c}{\textbf{Text-to-Video Retrieval}} &\multicolumn{4}{c}{\textbf{Video-to-Text Retrieval}} \\ 

\cmidrule(r){2-5} \cmidrule(r){6-9}
& R1 & R5 & R10 & SumR & R1 & R5 & R10 & SumR \\ 
\midrule
\rowcolor{gray!20}
\multicolumn{9}{@{}l}{\textbf{\textit{Vision-only:}}} \\
CLIP4Clip, Neucom22 \cite{clip4clip} & 46.4 & 72.1 & 82.0 & 200.5 & 45.4 & 73.4 & 82.4 & 201.2 \\
X-Pool, CVPR22 \cite{xpool} & 48.2 & 73.7 & 82.6 & 204.5 & 46.4 & 73.9 & 84.1 & 204.4 \\
TS2-Net, ECCV22 \cite{ts2net} & 49.4 & 75.6 & 85.3 & 210.3 & 46.6 & 75.9 & 84.9 & 207.4 \\
X-CLIP, MM22 \cite{xclip} & 49.3 & 75.8 & 84.8 & 209.9 & 48.9 & 76.3 & 85.4 & 210.6 \\
STAN, CVPR23 \cite{stan} & 50.0 & 75.2 & 84.1 & 209.3 & - & - & - & - \\
UATVR, ICCV23 \cite{uatvr} & 50.8 & 76.3 & 85.5 & 212.6 & 48.1 & 76.3 & 85.4 & 209.8 \\
UCoFiA, ICCV23 \cite{ucofia} & 49.8 & 74.6 & 83.5 & 207.9 & 49.1 & 77.0 & 83.8 & 209.9 \\
ProST, ICCV23 \cite{prost} & 49.5 & 75.0 & 84.0 & 208.5 & 48.0 & 75.9 & 85.2 & 209.1 \\
DGL, AAAI24 \cite{dgl} & 48.3 & 71.8 & 80.6 & 200.7 & 45.7 & 74.0 & 82.9 & 202.6 \\
EERCF, AAAI24 \cite{eercf} & 49.9 & 76.5 & 84.2 & 210.6 & 47.8 & 75.3 & 85.4 & 207.3 \\
TeachCLIP, CVPR24 \cite{teachclip} & 48.0 & 75.9 & 83.5 & 207.4 & - & - & - & - \\
TempMe, ICLR25 \cite{tempme} & 49.0 & 74.4 & 83.3 & 206.7 & 47.6 & 75.3 & 85.4 & 208.3 \\
DiscoVLA, CVPR25 \cite{discovla} & 50.5 & 75.6 & 83.8 & 209.9 & 49.2 & 76.0 & 84.7 & 209.9 \\
PIG, ICCV25 \cite{pig} & 51.2 & 75.1 & 84.5 & 210.8 & - & - & - & - \\
\midrule
\rowcolor{gray!20}
\multicolumn{9}{@{}l}{\textbf{\textit{Audiovisual:}}} \\
TEFAL, ICCV23 \cite{tefal} & 49.9 & 76.2 & 84.4 & 210.5 & - & - & - & - \\
AVIGATE, CVPR25 \cite{avigate} & 52.1 & 76.4 & 85.2 & 213.7 & 51.2 & 77.9 & 86.2 & 215.3 \\
\rowcolor{green!20}
SAVE & \textbf{54.5} & \textbf{80.3} & \textbf{87.4} & \textbf{222.2} & \textbf{51.4} & \textbf{80.1} & \textbf{87.0} & \textbf{218.5} \\
\bottomrule
\end{tabular}
}
\label{tab:vitb16}
\end{table}

\section{Qualitative Results}
\label{sup_sec:qualitative}
\begin{figure*}[!htbp]
  \centering
    \includegraphics[width=\linewidth]{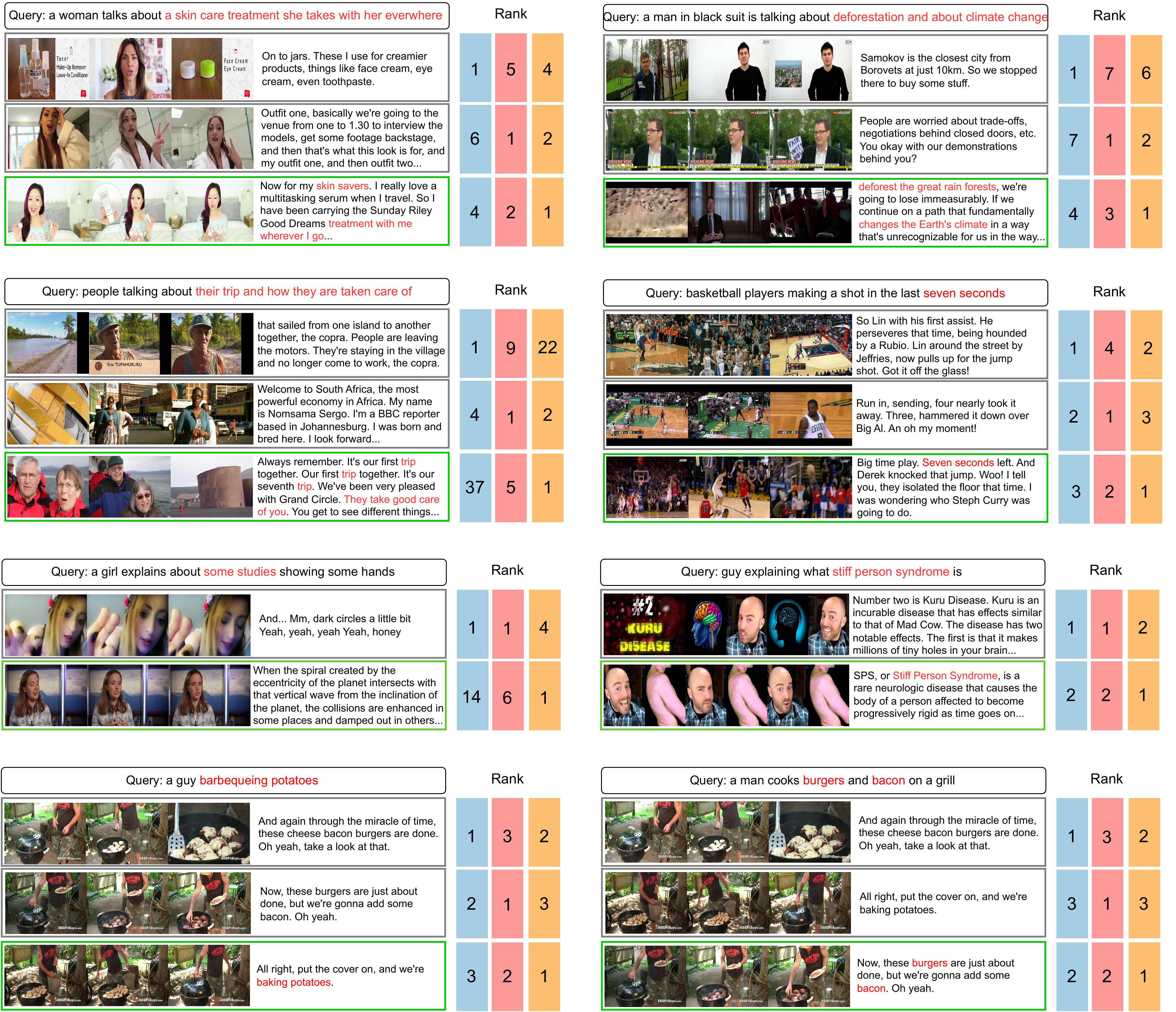}
    \caption{\textbf{Qualitative comparison} of the Top-1 retrieved videos from \colorbox[HTML]{A6CEE3}{\textbf{PIG}}, \colorbox[HTML]{FB9A99}{\textbf{AVIGATE}}, and \colorbox[HTML]{FDBF6F}{\textbf{SAVE}}, with the same color-coded blocks indicating how each model ranks these candidate videos. Benefiting from the speech branch, \colorbox[HTML]{FDBF6F}{\textbf{SAVE}} consistently assigns the ground-truth video (with green box) the best rank, showing clear advantages on speech-related queries. Best viewed on screen.}
  \label{fig:qualitative}
\end{figure*}

To further validate the effectiveness of our speech-aware video representation for text-to-video retrieval, we present qualitative visualizations in \cref{fig:qualitative}. SAVE consistently retrieves the correct video because the dedicated ASR-based speech branch effectively captures speech-related cues that PIG and AVIGATE fail to exploit.

\end{document}